\documentclass[runningheads]{llncs}
\usepackage{graphicx}
\usepackage{url}
\usepackage[hidelinks]{hyperref}
\usepackage{booktabs}
\begin{document}
\title{FPM: A Collection of Large-scale Foundation Pre-trained Language Models}
%
%
\author{
 Dezhou Shen}
%
\authorrunning{D. Shen.}
%
\institute{
\email{sdz15@tsinghua.org.cn}\\
}
\maketitle              
\begin{abstract}
Large-scale Transformer models have significantly promoted the recent development of natural language processing applications.
However, little effort has been made to unify the effective models.
In this paper, driven by providing a new set of baseline models in the future, we adopt various novel transformer architectures and launch a model set with the help of recent mainstream technologies.
We focus the discussions on optimizing the depth of the networks based on the existing powerful encode-decoder structures.
We show that by properly avoiding training defects such as non-convergence and degradation, scaling up off-the-shelf transformer architectures consistently delivers better performance.
To stimulate future research on large-scale language model pretraining, we present extensive results and detailed discussions on network performance improvements with respect to the network depth and confirm the existence of the optimal number of layers under specific tasks.
To the best of our knowledge, we provide the largest Chinese generative model and the largest Chinese encoding model. The BERT language models we trained on English datasets deliver a 14.45\% higher F1 score than the Turing-NLR.

\end{abstract}

\section{Introduction}

Thanks to the increasing availability of hardware and size of datasets, natural language processing has developed rapidly in recent years.
The massive resources of computation and data make it possible to train large-scale language models through self-supervised pre-training~\cite{qiu2020pre}. As the model size increases, the memory consumption can exceed the limit of the modern hardware. Fortunately, modern developments on distributed training have enabled models with hundreds of millions or even tens of billions of parameters to be trained on a large number of GPUs in parallel.
And now there are many techniques, such as layer normalization~\cite{ba2016layer} and residual connection~\cite{he2016identity} in Transformer layers~\cite{vaswani2017attention}, to eliminate model degradation when scaling up the training models.

However, little effort has been made to unify the effective models.
In this paper, driven by providing a new set of baseline models in the future, we adopt various novel transformer architectures and launch a model set with the help of recent mainstream technologies.

\par In summary, the contributions of this article are as follows:
\begin{itemize}
    \item We train a language model with 10.3 billion parameters, which to the best of our knowledge, is the largest Chinese generative model.
    \item We train a BERT language model with 495 million parameters, which to the best of our knowledge, is the largest Chinese encoding model.
    \item We train a GPT-2 language model with 6.4 billion parameters, which to the best of our knowledge, is the largest English generative model.
    \item We have trained a list of BERT language models, which to the best of our knowledge, exceeds the Turing-NLR's F1 score by 14.45\% on English datasets.
    \item We observe that the best result of Quora-Question-Pairs(QQP) from GLUE does not come along with the largest 90-layer-BERT-E-E model, and the 70-layer-BERT-E-L model achieves the state-of-the-art result with a 0.5\% accuracy improvement compared to the 90-layer-BERT-E-E model.
\end{itemize}


\section{Related Work}
\subsection{Terminology}

Pre-trained language models are classified into three categories: autoregressive language models (such as GPT), masked language models (such as BERT), and encoder-decoder models (such as BART, T5).
\par This paper uses three tokens, Chinese Pre-trained language Model (CPM)~\cite{zhang2020cpm}, English Pre-trained language Model (EPM)~\cite{zhang2020cpm}, and Generative Pre-Training (GPT)~\cite{radford2018improving}, to represent autoregressive language models.
We use BERT~\cite{devlin2019bert} to represent mask language models.
Moreover, we use three tokens, Cost-efficient Pre-trained language Models (CPM-2)~\cite{zhang2021cpm}, cost-efficient English Pre-trained Language Models (EPM-2)~\cite{zhang2021cpm} and Transformer~\cite{vaswani2017attention}, to represent encoder-decoder models.

\par Autoregressive language models predict the next word $x$, given all the previous words $x_1$, $x_2$, \ldots, and $x_{i-1}$. The training goal is to maximize the log likelihood $\sum_{i} m_{i} \log \left(P\left(x_{i} \mid x_{1}, \ldots, x_{i-1}, x_{i+1}, \ldots, x_{n}\right) ; \theta_{T}\right) $, where $\theta$ is the model parameter. Typical autoregressive language models include GPT, GPT-2~\cite{radford2020language} and GPT-3~\cite{brown2020language}.
\par Mask language models use the special token [MASK] to randomly select the word to be masked, or replace it with a random token. This architecture forces the model to collect bidirectional information when making predictions. Popular representations of MLM include BERT~\cite{devlin2019bert} and RoBERTa~\cite{liu2019roberta}. Specifically, MLMs such as BERT use Transformer encoder architecture. Like the autoregressive model, MLM stacks multiple Transformer encoder layers to learn increasing complex and meaningful representations, but when learning the representation of a specific token, it uses masked self-attention to focus on all others in the sequence token.
\par Encoder-decoder models use more flexible ``text input, text output" pattern, which learns to generate tokens $y_1$, \ldots, $y_n$ with the given input sequence $x_1$, \ldots, $x_m$. Representative models include BART~\cite{lewis2020bart} and T5~\cite{raffel2020exploring}.

\subsection{Pre-trained Models Review}

\par Autoregressive models include 2.6-billion-parameter CPM~\cite{zhang2020cpm}, 2.7-billion-parameter GPT-Neo, 6-billion-parameter GPT-J and 175-billion-parameter GPT-3~\cite{brown2020language}.
\par Mask language models include 330-million-parameter Roberta model,  27-billion-parameter PLUG~\cite{lin2021m6} model, 110-billion-parameter Pangu~\cite{zeng2021pangu} model, and 1.6-billion-parameter CTRL model.
\par Encoder-decoder models include 11-billion-parameter CPM-2~\cite{zhang2021cpm} model.
\par The Table-\ref{tab:modelreview} shows the popular models performance on GLUE-QQP test set. In the Section 4.3, we will compare the performance of ours with existing popular models.

\begin{table}[]
    \begin{center}
                \caption{Models performance on GLUE-QQP test set.}
                        \label{tab:modelreview}
\begin{tabular}{lrr}
    \toprule
\textbf{model}                      & \textbf{QQP-acc.} & \textbf{QQP-f1} \\
    \midrule
BiLSTM+Attn, ELMo~\cite{Peters:2018} & 86.5\%                   & 66.1\%                 \\
BERT-BASE~\cite{devlin2019bert}       & -                       & 71.2\%                 \\
BERT-LARGE~\cite{devlin2019bert}      & -                       & 72.1\%                 \\
GPT~\cite{radford2018improving}       & -                       & 70.3\%                 \\
ERINE2.0-BASE~\cite{sun2020ernie}      & 89.8\%                 & 73.2\%                  \\
ERINE2.0-LARGE~\cite{sun2020ernie}     & 90.1\%                 & 73.8\%                   \\
XLNet~\cite{yang2019xlnet}             & 90.4\%                 &  -                       \\
ELECTRA~\cite{clark2020electra}        &  90.8\%                & -                       \\
ERINE~\cite{sun2019ernie}              & 90.9\%                 & 75.2\%                   \\
Turing NLR v5~\cite{glue2019leader}    & 91.1\%                 & 76.4\%                   \\
       \bottomrule
\end{tabular}
        \end{center}
\end{table}

\section{Architecture}

Based on the GPT, BERT, Transformer network structure, we extend the GPT, BERT, and Transformer network structure to different numbers of the network layers. We study the performance differences among the models by only changing the number of layers.

\subsection{GPT}
In experiment of the original GPT article, Radford et al. use multi-layer transformer decoders in the language model, which is a variant of Transformer. It applies a multi-head self-attention operation to the input tokens, and a position feed-forward layer to generate the output distribution on target tokens.

\par We study the performance impact of the model depth by testing the CPM model with 36, 64, 128 layers and the EPM model with 36, 50, 64 and 80 layers. All other hyper-parameters are identical with the original models. We expanded network layers number from 32 layers to 36, 64, and 128  for the CPM model, and we have expanded layers from 32 to 36, 50, 64, and 80 for the EPM model.

\subsection{BERT}
\par BERT is a pre-trained Transformer network that sets up the latest state-of-the-art results for various natural language processing tasks, including question answering, sentence classification, and sentence pair regression.
\par The BERT-Large model mentioned in the paper by Devlin et al. consists of 24 self-attention layers. In this study, we extend the original network from 24 layers to 50, 60, 70, 80, and 90 layers, with all other hyper-parameters unchanged. The performance impact of the model depth is also discussed in Table-\ref{tab:qqpeval}.

\subsection{Transformer}
\par CPM-2 is a standard Transformer model that combines a bidirectional encoder and a directional decoder.
In order to reduce memory consumption and accelerate pre-training, Zhang et al. use mixed-precision training, gradient checkpointing, and zero stage optimization~\cite{rajbhandari2020zero}.

\par Most powerful neural sequence transduction models have an encoder-decoder structure.
Transformer~\cite{vaswani2017attention} follows this overall architecture, and uses stacked self-attention and point-by-point fully connected encoder and decoder layers.

\par We reduce the original CPM-2 model from 48 layers of the Transformer unit to 12 and 24 layers. All other hyper-parameters in our model are consistent with the CPM-2 model in the original paper. Moreover, we train the 12-layer EPM-2 model on English datasets.

\section{Experiments}
\subsection{Setup}

In this work, we focus on three models: GPT-2, BERT, CPM-2, a language model based on a left-to-right generative transformer, a bidirectional transformer model based on masking, an encoder-decoder language model, respectively.

\subsubsection{Training Datasets}
\par A dataset should be as large as possible with high quality first, the classification should be as even as possible, and the content of the data should be as clean as possible. This work only focuses on model training, and uses existing  large-scale datasets, the diversified English Pile dataset and the Chinese Wudao dataset.
\paragraph{GPT-English} word segmentation uses GPT-2 English vocabulary, containing 30,000 symbols. OpenAI team used 40GB of text and 8 million documents collected by WebText corpus. WebText is a web page curated and filtered by OpenAI humans. All outbound links with a rating of at least 3 karma are crawled from Reddit. The generated dataset WebText contains a text subset of 45 million links. In the cleanup phase, links created after December 2017 were removed, and after deduplication and some heuristic-based cleanup, slightly more than 8 million documents were left, a total of 40GB of text, and all Wikipedia documents were deleted.
\paragraph{GPT-Chinese} word segmentation uses CPM Chinese vocabulary, containing 30,000 symbols. Zhang et al. use 100G multi-category texts, including encyclopedias, news, novels, and questions-answers. CPM Chinese vocabulary uses the unigram language model to build a new sub-word vocabulary based on a sub-word corpus. Since the length of input sequence is usually greater than the length of a single document, different documents are connected by adding the ``end-of-document" symbol after each document. To make full use of input length, A new sub-word vocabulary is constructed in the vocabulary construction process, including commonly used words and characters. Considering that original BERT word segmentation will introduce an additional splitter between words, Zhang et al. set up a special token as a splitter to make the sub-word process reversible.
\paragraph{Transformer-Chinese} word segmentation uses CPM-2 Chinese vocabulary. It contains 26,240 symbols and is trained by Zhang et al. using 2.3TB of cleaned Chinese data. CPM-2's vocabulary is modified based on Chinese byte-pair encoding (BPE). Original BPE inserts many redundant space marks ``\_" in word segmentation sequence. Zhang et al. replace sentence segmentation device with a combination of word segmentation device and the stammering word segmentation, and deletes it Inserted spaces. Since it does not matter whether symbols in vocabulary appear at the beginning of the word, tags like ``happy" (happy) and ``\_happy" (\_happy) are merged into a single token ``happy" to simplify the vocabulary.
\paragraph{Transformer-English} word segmentation, using CPM-2 English vocabulary, contains 29,752 symbols, which is trained by Zhang et al. using 300GB of cleaned English data. CPM-2's vocabulary is modified based on English BPE. Original BPE inserts many redundant space tokens ``\_" in word segmentation sequence. Zhang et al. replaced the sentence tokenizer with word tokenizer combined with nltk word segmentation, and deleted inserted spaces. English data comes from multiple fields, including encyclopedias, novels, Q\&A, scientific literature, e-books, news, and reviews.
\par We use BERT's English vocabulary for BERT-English word segmentation, which contains 30,000 symbols. It was trained by Devil et al. using WordPiece embedding on BooksCorpus (800M words) and English Wikipedia (2500M words).
\par For BERT-Chinese word segmentation, we use BERT's Chinese vocabulary, which contains 21,128 symbols and is trained by Cui et al. using 13.6 million lines of Wikidata in Chinese. In the process of generating BERT vocabulary, Cui et al. downloaded the latest Wikipediadump and used WikiExtractor.py to preprocess it according to the recommendations of Devlin et al., resulting in 1,307 extracted files, and used language technology platform for Chinese word segmentation.
\par We use an 800GB corpus from Pile. Due to the limitation of computing resources, not all partitions was selected in the process. We use only four blocks of 02, 03, 04, 17 about 213G data. Two types of the corpus, StackExchange and Github, which contain many impurities, are included, and the entire dataset is about 200G.
\par We use 3TB Chinese corpus collected by Yuan et al. In the process of using Wudao corpus, due to the limitation of computing resources, we use 200G corpus. Wudao dataset uses 3 billion web pages as an original data source to extract text content from web pages with high text density. Evaluate the quality of each data source before extracting the text, and ignore web pages with text density below 70\%.
\par The Chinese dialogue data from STC-680M corpus dataset, contains approximately 4.4 million conversations from Weibo. To build this million-scale dataset, they first grab hundreds of millions of response pairs and then filter out potential responses by deleting trivial answers such as ``wow", Advertisements, and delete the content after the first 30 responses to keep the theme consistent with cleaning up original data.


\subsubsection{Training Optimization and Hyperparameters}
\par To effectively train the model, in some experiments, mixed-precision training and dynamic loss scaling are removed to use a tensor core of A100. However, in some experiments, training fails to converge due to accuracy, and we remove the mixed-precision training method.
First initialize weights with a simple normal distribution. Then scale the weights immediately before the residual layer, where N is the number of transformer layers composed of self-attention and Multilayer Perceptron (MLP) blocks.
For optimizer, use Adam with a weight decay of 0.01.
In addition, a 1.0 global gradient norm crop is used to improve the stability of training large models. In all cases, use a dropout of 0.1. Finally, to better manage the memory footprint, activation checkpoints are used after each transformer layer.
\par For GPT-2 model, we make all training experiments using 1024 symbol sequences, and batch size is 512, and 300k iterations are performed. We preheat 1.5e-4 learning rate for 3k iterations, and then a single-loop cosine decay is performed in the remaining iterations. Stop attenuation at the minimum learning rate of 1e-5.
\par For BERT, we follow the training process described in original paper. Using original BERT dictionary, vocabulary size is 30,522. In addition, follow suggested sentence order prediction to reposition the next sentence prediction head and use the whole word n-gram mask. For all cases, set batch size to 1024, and use a learning rate of 1.0e-4, warm up in 10,000 iterations and decay linearly in remaining iterations. Other training parameters remain unchanged.

\begin{table}[]
    \begin{center}
            \caption{Corpus language, models parameters and layers for FPM models.}
                    \label{tab:corpusparam}
\begin{tabular}{llllllll}
    \toprule
\textbf{model}      & \textbf{corpus-size} & \textbf{language}   & \textbf{$n_{parameters}$} & \textbf{$n_{layers}$} & \textbf{$d_{layer}$} & \textbf{$n_{heads}$} & \textbf{$d_{head}$} \\
\midrule
\textbf{BERT-C}      & 200G                 & CN                & 330M                  & 24               & 1024                  & 16             & 64                   \\
\textbf{BERT-E-S}    & 200G                 & EN                & 687.5M                & 50               & 1024                  & 16             & 64                   \\
\textbf{BERT-E-M}    & 200G                 & EN                & 825M                  & 60               & 1024                  & 16             & 64                   \\
\textbf{BERT-E-L}    & 200G                 & EN                & 962.5M                & 70               & 1024                  & 16             & 64                   \\
\textbf{BERT-E-X}    & 200G                 & EN                & 1.1B                  & 80               & 1024                  & 16             & 64                   \\
\textbf{BERT-E-E}    & 200G                 & EN                & 1.24B                 & 90               & 1024                  & 16             & 64                   \\
\textbf{BERT-X-CN-S} & 200G                 & CN                & 495M                  & 36               & 1024                  & 16             & 64                   \\
\textbf{BERT-X-EN-S} & 200G                 & EN                & 495M                  & 36               & 1024                  & 16             & 64                   \\
\textbf{BERT-X-EN-M} & 200G                 & EN                & 687.5M                & 48               & 1024                  & 16             & 64                   \\
\textbf{CPM-X-S}     & 200G                 & CN                & 2.9B                  & 36               & 2560                  & 32             & 80                   \\
\textbf{CPM-X-M}     & 200G                 & CN                & 5.1B                  & 64               & 2560                  & 32             & 80                   \\
\textbf{CPM-X-L}     & 200G                 & CN                & 10.3B                 & 128              & 2560                  & 32             & 80                   \\
\textbf{CPM-X-EVA}     & 684M                 & CN                & 2.9B                  & 36               & 2560                  & 32             & 80                   \\
\textbf{EPM-X-S}     & 200G                 & EN                & 2.9B                  & 36               & 2560                  & 32             & 80                   \\
\textbf{EPM-X-M}     & 200G                 & EN                & 4B                    & 50               & 2560                  & 32             & 80                   \\
\textbf{EPM-X-L}     & 200G                 & EN                & 5.1B                  & 64               & 2560                  & 32             & 80                   \\
\textbf{EPM-X-X}     & 200G                 & EN                & 6.4B                  & 80               & 2560                  & 32             & 80                   \\
\textbf{CPM-2-X-S}   & 200G                 & CN                & 2.9B                  & 12               & 4096                  & 64             & 64                   \\
\textbf{CPM-2-X-M}   & 200G                 & CN                & 5.6B                  & 24               & 4096                  & 64             & 64                   \\
\textbf{EPM-2-X-S}   & 200G                 & EN                & 2.9B                  & 12               & 4096                  & 64             & 64                    \\
 \bottomrule
\end{tabular}
        \end{center}
\end{table}

\subsection{Model Details}

The infrastructure is optimized for multi-node deep learning applications. All experiments use up to 2 DGX servers (a total of 16 A100 SXM3 40GB GPUs). Through NVSwitch, we achieved a bandwidth of 300GB/sec between GPUs in servers and achieved a bandwidth of 10GB/sec between servers that use 1 InfiniBand adapter per server.

\par We describe the most significant models as follows:
\begin{itemize}
\item Based on GPT-2, we trained a 10.3-billion-parameter model on Chinese datasets and we trained a 2.9-billion-parameter model on a dialogue corpus. We trained a BERT model with 495 million parameters on Chinese datasets. Moreover, we trained a Transformer model with 5.6 billion parameters on Chinese datasets.
\item We apply the corresponding training work for English. Using the GPT-2 model, we trained a model with 6.4 billion parameters on English datasets.  We trained a BERT model with 1.24 billion parameters on English datasets and we trained a language model with a 688 million parameter on one GPU. We trained a Transformer model with 2.9 billion parameters on English datasets.
\end{itemize}

\subsubsection{EPM-X}
Based on GPT-2 model, we have built an encoder-decoder language model EPM-2-X, using GPT-English tokenizer, and trained a language model with 2.9 billion parameters. It has a 12-layer network structure, 6 encoding layers, and 6 decoding layers.

\subsubsection{EPM-2-X}
Based on Transformer model, we have built an encoder-decoder language model EPM-2-X, using Transformer-English tokenizer, and trained a language model with 2.9 billion parameters. It has a 12-layer network structure, 6 encoding layers, and 6 decoding layers.

\subsubsection{BERT-E}
Based on BERT model, we built encoding language model BERT-E. Using BERT-English word tokenizer, we trained five models with different layers. The largest model is a language model with 1.24 billion parameters. It has 90 layers of Transformer encoding layers. The network is cascaded, named BERT-E-E. The second E here means Extreme.

\subsubsection{BERT-X-EN}
Based on BERT model, we have built encoding language model BERT-X-EN. Using BERT-English word segmentation, we trained two models with different layers. The largest model is 690 million parameters, includes 48 layers of Transformer encoding layer stack, named BERT-X-EN-M.

\subsubsection{CPM-X}
Based on GPT-2 model, we have built a generative language model CPM-X, using GPT-Chinese word segmentation, and trained 3 models with different sizes. The largest model is a 10.3 billion parameter language model, which stacks 128 layers Transformer decoding layer, named CPM-X-L.
Based on GPT-2 model, we have built a generative language model CPM-X-EVA, using GPT-Chinese word segmentation, using STC dialogue data, a language model of 2.9 billion parameters, which stacked 36 layers of Transformer decoding layer, named CPM-X-EVA.

\subsubsection{CPM-2-X}
Based on Transformer model, we built Transformer language model CPM-2-X. We trained 2 models, the largest of which uses Transformer-Chinese word segmentation, trained a 5.6 billion parameter language model. It has a 24-layer network structure, 12 encoding layers and 12 decoding layers, named CPM-2-X-M.

\subsubsection{BERT-C}
Based on BERT model, we built encoding language model BERT-C. Using a BERT-Chinese word tokenizer, we trained a language model with 330 million parameters. It has a 24-layer Transformer encoding layer network cascaded, named BERT-C.

\subsubsection{BERT-X-CN}
Based on BERT model, we built encoding language model BERT-X-CN. Using a BERT-Chinese word tokenizer, we trained a language model with 495 million parameters, including 36 layers of Transformer encoding layers, named BERT-X-CN-S.


\subsection{Evaluation}

In evaluation stage, we evaluate five BERT-E models on the QQP classification task from GLUE and evaluate BERT-C model on the TNEWS classification task from CLUE.

\begin{table}[]
    \begin{center}
            \caption{Evaluation results on QQP dev set.}
                        \label{tab:qqpeval}
\begin{tabular}{lllll}
    \toprule
\textbf{model} &  \textbf{precision} & \textbf{recall} & \textbf{f1} & \textbf{acc.} \\
  \midrule
\textmd{BERT-E-S}             & 89.84\%            & 90.42\%         & 90.11\%     & 90.73\%           \\
\textmd{BERT-E-M}             & 89.84\%            & 90.43\%         & 90.12\%     & 90.73\%           \\
\textmd{BERT-E-L}             & \textbf{90.55}\%            & \textbf{91.19}\%         & \textbf{90.85}\%     & \textbf{91.42}\%          \\
\textmd{BERT-E-X}             & 89.99\%            & 90.39\%         & 90.18\%     & 90.82\%           \\
\textmd{BERT-E-E}             & 90.00\%            & 90.68\%         & 90.31\%     & 90.91\%           \\
\bottomrule
\end{tabular}
        \end{center}
\end{table}

\subsubsection{GLUE-QQP}
We use an Nvidia 3090 GPU to finetune five BERT-E models on QQP dataset. The QQP task has 363k, 40k and 390k records in the training, dev and test set.
From Table-\ref{tab:qqpeval},
BERT-E-L model has 70 layers, 1024 hidden dimensions, 16 attention heads, each head contains 64 hidden layers, and 962.5 million parameters.
In QQP classification task evaluated by GLUE, after 45 hours of fine-tuning 270k steps on 109M corpus, the accuracy of 91.42\% exceeded an accuracy of BERT-Large of 72.1\%, an increase of 19.32\%.
From the precision and recall evaluation, compared with Turing-NLR's F1, the first place in GLUE evaluation of 76.4\%, our BERT-E-L model F1 score gained an increase of 14.45\%.

\subsubsection{CLUE-TNEWS}
We use an Nvidia 3090 GPU to finetune BERT-C model on TNEWS dataset.
In TNEWS classification task evaluated by CLUE, after fine-tuning the 9.7M corpus for 8 hours 70,000 steps, the accuracy of 65.86\% surpassed 59.46\% of ALBERT-xxlarge, an increase of 6.4\%.

\begin{table}[]
    \begin{center}
            \caption{Cost for FPM models.}
                    \label{tab:corpuscost}
\begin{tabular}{lrrrr}
\toprule
\textbf{model}       & \textbf{time} & \textbf{step} & \textbf{gpus} & \textbf{flops(Eflops)} \\
\midrule
\textmd{BERT-C}      & 81h           & 750K         & 8          & 727   \\
\textmd{BERT-E-S}    & 34h           & 500K         & 8          & 305   \\
\textmd{BERT-E-M}    & 38h51m        & 500K         & 8          & 349   \\
\textmd{BERT-E-L}    & 45h38m        & 500K         & 8          & 410   \\
\textmd{BERT-E-X}    & 54h24m        & 500K         & 8          & 488   \\
\textmd{BERT-E-E}    & 63h           & 500K         & 8          & 566   \\
\textmd{BERT-X-CN-S} & 96h           & 880K         & 1          & 107   \\
\textmd{BERT-X-EN-S} & 315h          & 2.8M         & 1          & 353   \\
\textmd{BERT-X-EN-M} & 315h          & 2.8M         & 1          & 353   \\
\textmd{CPM-X-S}     & 3h            & 10K          & 8          & 27  \\
\textmd{CPM-X-M}     & 12h           & 60K          & 8          & 108   \\
\textmd{CPM-X-L}     & 24h           & 100K         & 8          & 216   \\
\textmd{CPM-X-EVA}       & 25h           & 160K         & 2          & 54    \\
\textmd{EPM-X-S}     & 20h           & 320K         & 4          & 92    \\
\textmd{EPM-X-M}     & 24h           & 320K         & 4          & 110   \\
\textmd{EPM-X-L}     & 27h           & 320K         & 4          & 121   \\
\textmd{EPM-X-X}     & 30h           & 320K         & 4          & 135   \\
\textmd{CPM-2-X-S}   & 60h           & 200K         & 2          & 135   \\
\textmd{CPM-2-X-M}   & 138h          & 80K          & 8          & 1240  \\
\textmd{EPM-2-X-S}   & 110h          & 200K         & 2          & 247  \\
\bottomrule
\end{tabular}
        \end{center}
\end{table}

\section{Discussion}

We will discuss language, scale, network configuration, and cost for our training models. Furthermore, we will discuss the optimal layer number for the GLUE-QQP task.

\subsection{Language, Scale and Architecture}

\par From Table-\ref{tab:corpusparam}, this work has trained many Chinese and English corpora and conducted detailed training.

\par The models with the least number of layers are CPM-2-X and EPM-2-X, with only 12 layers. Model with the most significant number of layers is CPM-X, with 128 layers trained. From perspective of the difficulty of training, GPT structure is easier to cascade the number of layers. While Transformer structure is not easy to increase the number of layers, it is easier to reach the upper limit of hardware memory.
From a perspective of model parameters, model with the least amount of parameters is BERT-C, with only 330 million, and the most significant parameter is CPM-X-L, with parameters reaching 10.3 billion.

\par For BERT, GPT, and Transformer, our network structure parameters used in this work are fixed, and we use parameters in the original paper without significant modifications.

\subsection{Cost}
We observe that in terms of time consumption, the shortest training time is CPM-X-S, and the longest time is CPM-X-EN from Table-\ref{tab:corpuscost}. The least training step number is CPM-X-S, only 100,000 steps. Moreover, the largest is CPM-X-EN with 2.8 million steps. From the perspective of computing power, the least computing resource is CPM-X-S, which uses 27 Eflops, and the most computing power is CPM-2-X-M, which uses 1,240 EFlops.

\subsection{Optimal Layers}
From Table-\ref{tab:qqpeval}, we observe that the best result does not come with the largest 90-layer-BERT-E-E model, the 70-layer-BERT-E-L model achieved the state-of-the-art result with about 0.5\% gain to the 90-layer-BERT-E-E model.
We can conclude that for the GLUE-QQP task, the optimal layer number for BERT models is 70. We conclude the result from the comparison with the current GPU hardware limits.
In the future, the optimal layer number for the GLUE-QQP task might change with the help of advanced GPU servers.

\section{Conclusion}

\par We focus the discussions on optimizing the depth of the networks based on the existing powerful encode-decoder structures.
We observe that the best result of QQP from GLUE does not come with the largest 90-layer-BERT-E-E model, the 70-layer-BERT-E-L model achieved the state-of-the-art result with about 0.5\% gain to the 90-layer-BERT-E-E model.

\section{Future Work}

With the development of hardware and software, we will try more efficient model tricks for the deeper network layers and train more improved Chinese and English models based on BERT and GPT in future. In addition, we will explore the optimal structure in models for different numbers of model layers in accuracy.

%
%
%
%
\bibliographystyle{splncs04}
\bibliography{founder-pm-arxiv}
\end{document}